\documentclass[runningheads]{llncs}
\usepackage{tabularx, booktabs}
\usepackage{threeparttable}
\newcommand{\corr}{\textsuperscript{\dag}}
\usepackage{caption}
\newlength{\origtabcolsep}
\usepackage{amsmath,amssymb, amsfonts, stmaryrd}
\usepackage{hyperref}
\usepackage{multirow}
\usepackage{marvosym} 

\DeclareRobustCommand{\corr}{\textsuperscript{\Letter}}
\usepackage{wrapfig}
\usepackage{newclude}
\usepackage{bm}
\usepackage{xcolor}   
\usepackage{amssymb}  
\usepackage{bbm}
\usepackage{makecell} 
\usepackage{booktabs}
\usepackage{array} 
\usepackage{xcolor}
\usepackage{pifont}
\usepackage{graphicx}
\usepackage{colortbl}

\newcommand{\metbf}[1]{\scalebox{0.85}{\textbf{#1}}}

\begin{document}
\title{BrainAnytime: Anatomy-Aware Cross-Modal Pretraining for Brain Image Analysis with Arbitrary Modality Availability}
\titlerunning{BrainAnytime: Anatomy-Aware Pretraining for Multi-modal Brain Imaging}
%
%

\author{Guangqian Yang\inst{1}* \and Tong Ding\inst{1}* \and Wenlong Hou\inst{1} \and Yue Xun\inst{1} \and Ye Du\inst{1} \and Qian Niu\inst{2}\corr \and Shujun Wang\inst{1,3}\corr \and for the Alzheimer’s Disease Neuroimaging Initiative}

\institute{Department of Biomedical Engineering, The Hong Kong Polytechnic University, Hong Kong SAR, China
\and Department of Technology Management for Innovation, The University of Tokyo, Japan
\and Department of Data Science and Artificial Intelligence, The Hong Kong Polytechnic University, Hong Kong SAR, China \\
\email{qian.niu@weblab.t.u-tokyo.ac.jp, shu-jun.wang@polyu.edu.hk}\\
\Letter~Corresponding authors.\\
*These authors contributed equally to this work.}

\authorrunning{G. Yang et al.}

\maketitle              
%
\begin{abstract}


Clinical diagnostic workups typically follow a modality escalation pathway: after initial clinical evaluation, clinicians begin with routine structural imaging (\textit{e.g.}, MRI), selectively add sequences such as FLAIR or T2 to refine the differential, and reserve molecular imaging (\textit{e.g.}, amyloid-PET) for cases that remain uncertain after standard evaluation. Consequently, patients are observed with heterogeneous and often incomplete modality subsets. However, most current AI models assume fixed data modalities as the model inputs.
In this paper, we present BrainAnytime, a unified pretraining framework pretrained on \textbf{34,899} 3D brain scans from five datasets that support brain image analysis under arbitrary modality availability spanning multi-sequence MRI and amyloid-PET. A single model accepts whatever imaging is available, from a lone T1 scan to a full multimodal workup. Pretraining learns structural-molecular correspondences between MRI and PET via cross-modal distillation (RCMD) and prioritizes disease-vulnerable anatomy via atlas-guided curriculum masking (PACM), all within a shared 3D masked autoencoder (Multi-MAE3D). Across \textbf{four} downstream tasks and \textbf{five} clinically motivated modality settings, BrainAnytime largely outperforms modality-specific models, missing-modality baselines, and large-scale brain MRI pretrained foundation models on most modality settings. Notably, it surpasses the strongest missing-modality baselines with relative improvements of \textbf{6.2\%} and \textbf{7.0\%} in average accuracy on CN vs. AD and CN vs. MCI classification, respectively. Code is available at \url{https://github.com/SDH-Lab/BrainAnytime}.

\end{abstract}
\section{Introduction}
When a patient presents with suspected cognitive decline, the diagnostic workup typically begins with clinical history review and neuropsychological screening~\cite{jack2024revised}. Once these initial evaluations warrant neuroimaging, clinicians rarely order every modality upfront. Instead, imaging follows a staged escalation of evidence: T1-weighted MRI is acquired first, additional MRI sequences
(\textit{e.g.}, FLAIR and T2) are added as needed, and amyloid-PET, which measures \(\beta\)-amyloid plaque burden and serves as a core biomarker of AD pathology, is often reserved for cases where structural imaging alone is insufficient~\cite{chen2025multi}. We refer to this sequential acquisition logic as the \textit{clinical modality escalation pathway}. Consequently, modality availability is inherently incomplete and stage dependent: in major AD cohorts, only a small fraction of subjects have a complete set of all imaging modalities (Fig.~\ref{fig:framework} and Table~\ref{tab:dataset_modality_stats}), while most are observed through partial and heterogeneous modality subsets.

These practice patterns motivate a different pretraining paradigm.
Rather than training separate models for fixed input configurations,
we seek a unified pretraining model that supports brain image analysis under \textit{arbitrary modality availability} spanning multi-sequence MRI and amyloid-PET.
Such a framework should satisfy three requirements.
\textbf{(i) Modality-flexible inference.}
A single model should accept any subset of available modalities at test time without architectural switching or retraining.
\textbf{(ii) Cross-modal structural-molecular correspondence.}
Pretraining should encourage shared representations that capture correspondences between structural MRI and amyloid-PET, instead of learning each modality in isolation~\cite{chetelat2020amyloid}.
\textbf{(iii) Anatomy-aware learning.}
Because AD exhibits selective regional vulnerability~\cite{hu2025anatomy}, pretraining should emphasize disease-salient anatomy rather than treating all locations uniformly.

However, existing brain imaging pretraining paradigms do not meet these requirements in the AD setting.
Recent large-scale self-supervised models pretrained on brain neuroimaging show strong transfer across downstream tasks~\cite{tak2026generalizable, DENG2025112595, rui2025multi, 11023095, ding2025denseformer, yin2025unicross, erdur2025multimae, zhang2025foundation}.
Despite this progress, three limitations remain.
First, most are \textbf{MRI-only} and therefore do not learn an explicit bridge to amyloid-PET, even though PET is a key source of pathological evidence along the escalation pathway.
Second, pretraining and deployment commonly assume a \textbf{fixed modality} (or a fixed combination), mismatching real-world, stage-dependent missingness and hindering inference under arbitrary modality subsets.
Third, standard MAE pretraining~\cite{he2022masked} relies on \textbf{spatially uniform masking}, providing no mechanism to prioritize anatomically vulnerable regions that are most informative for AD.
These gaps motivate a cross-modal, modality-flexible, and anatomy-aware pretraining framework tailored to clinically realistic modality availability.

We present \textbf{BrainAnytime}, an anatomy-aware cross-modal pretraining framework for brain image analysis with \textit{arbitrary modality availability}. BrainAnytime is pretrained on \textbf{34,899} 3D brain scans from \textbf{five} large-scale datasets spanning multi-sequence MRI and amyloid-PET, by randomly sampling modality subsets to mimic clinical missingness. BrainAnytime integrates three components: \textbf{Multi-MAE3D}, a 3D multi-modal masked autoencoder with a shared Transformer encoder for any modality subset; \textbf{Reciprocal Cross-Modal Distillation (RCMD)}, an EMA-teacher distillation objective that aligns MRI and PET representations; and \textbf{Pathology-Aware Curriculum Masking (PACM)}, an atlas-guided curriculum masking strategy that emphasizes AD-relevant neuroanatomy during reconstruction. Together, these designs yield a unified pretrained model that is cross-modal, robust to missing modalities, and explicitly anatomy-aware.
\section{Method}
\begin{figure}[!t]
\centerline{\includegraphics[width=1\columnwidth]{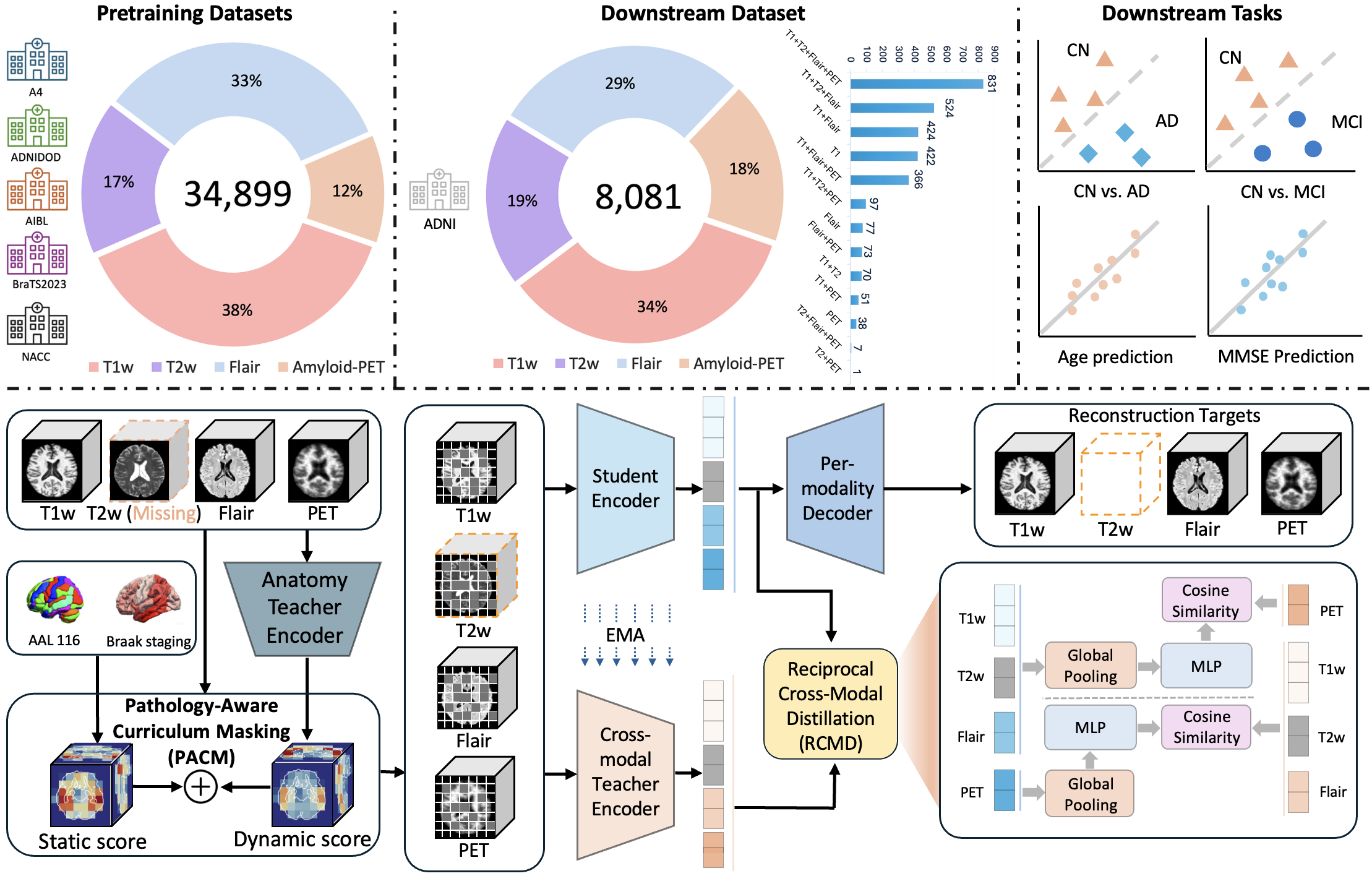}}

\caption{Overall framework of BrainAnytime.
} 
\label{fig:framework}
\end{figure}

We propose \textit{Anatomy-Aware Cross-Modal Pretraining}, a self-supervised framework for learning unified representations from heterogeneous 3D neuroimaging data with arbitrary missing modalities.
As shown in Fig.~\ref{fig:framework}, our framework consists of three components:
(i)~a 3D multi-modal masked autoencoder (\textbf{Multi-MAE3D}) that encodes any subset of modalities with a single Transformer;
(ii)~\textbf{Reciprocal Cross-Modal Distillation (RCMD)} that learns MRI--PET correspondences via an EMA teacher;
and (iii)~\textbf{Pathology-Aware Curriculum Masking (PACM)} that progressively focuses reconstruction on disease-relevant brain regions.

\subsection{Multi-MAE3D: 3D Multi-Modal Masked Autoencoder}
\label{sec:multimae}

Given a set of 3D volumes $\{\mathbf{x}_m\}_{m \in \mathcal{M}}$ where $\mathcal{M} \subseteq \{\text{T1, T2, FLAIR, PET}\}$ denotes the available modalities, each volume $\mathbf{x}_m \in \mathbb{R}^{D \times H \times W}$ is partitioned into $N$ non-overlapping patches of size $p^3$.

\noindent\textbf{Per-modality tokenization.}
Each modality has a modality-specific patch embedding $f_m$ with a 3D convolutional layer, that maps raw patches to $d$-dimensional tokens.
Following PACM (Sec.~\ref{sec:masking}), only a subset of visible tokens per modality is retained. Missing modalities are fully masked, and the visible-token budget is 
redistributed among observed modalities via 
$\text{Dir}(\alpha)$~\cite{erdur2025multimae} under a global mask 
ratio~$r$, keeping the total token count constant regardless of 
modality availability.

\noindent\textbf{Shared encoder.}
The visible tokens from all available modalities are concatenated with a learnable \texttt{[CLS]} token, augmented with 3D sinusoidal positional embeddings, and processed by a shared student encoder.
An attention mask blocks tokens from missing modalities (set to $-\infty$), ensuring the encoder gracefully handles any modality subset at both training and inference time.

\noindent\textbf{Per-modality decoders.}
Each modality has a lightweight decoder that takes the encoded visible tokens and learnable mask tokens as input.
Cross-attention between mask tokens and encoder output is followed by self-attention blocks.
The decoder reconstructs \emph{all} $N$ patches per modality; the training loss is computed only on masked patches using mean squared error with per-patch normalization~\cite{he2022masked}:
\begin{equation}
\mathcal{L}_{\text{MAE}} = \frac{1}{|\mathcal{M}|}\sum_{m \in \mathcal{M}} \frac{1}{|\mathcal{S}_m|} \sum_{i \in \mathcal{S}_m} \left\| \hat{\mathbf{p}}_i^m - \bar{\mathbf{p}}_i^m \right\|_2^2,
\label{eq:mae}
\end{equation}
where $\mathcal{S}_m$ is the set of masked patch indices for modality $m$, $\hat{\mathbf{p}}_i^m$ is the predicted patch, and $\bar{\mathbf{p}}_i^m$ is the per-patch normalized ground truth.

\subsection{Reciprocal Cross-Modal Distillation (RCMD)}
\label{sec:crossmodal}

To explicitly learn MRI--PET correspondences, we introduce RCMD, a cross-modal prediction objective based on an exponential moving average (EMA) cross-modal teacher.
The teacher shares the encoder architecture and is updated as $\theta_T \leftarrow \mu\,\theta_T + (1{-}\mu)\,\theta_S$, where the momentum $\mu$ follows a cosine schedule from $\mu_0$ to $1.0$. For paired samples containing both MRI and PET, we compute group-level 
representations by average-pooling the encoded tokens within two groups: $\mathbf{z}_{\text{MRI}}$ (T1+T2+FLAIR tokens) and $\mathbf{z}_{\text{PET}}$ (PET tokens), from both the student and teacher encoders.
Two symmetric MLP predictors $g_{\text{M}\to\text{P}}$ and $g_{\text{P}\to\text{M}}$ are trained to predict cross-modal teacher features:
\begin{equation}
\mathcal{L}_{\text{RCMD}} = 1 - \frac{1}{2}\!\left[\cos\!\big(g_{\text{M}\to\text{P}}(\mathbf{z}_{\text{MRI}}^s),\, \mathbf{z}_{\text{PET}}^t\big) + \cos\!\big(g_{\text{P}\to\text{M}}(\mathbf{z}_{\text{PET}}^s),\, \mathbf{z}_{\text{MRI}}^t\big)\right],
\label{eq:rcmd}
\end{equation}
where superscripts $s$ and $t$ denote the student and teacher, respectively, and all representations are $\ell_2$-normalized; \(cos(\cdot)\)  denotes cosine similarity.

\subsection{Pathology-Aware Curriculum Masking (PACM)}
\label{sec:masking}
  Standard MAE treats all spatial locations equally during masking.
  In neuroimaging, however, certain regions carry greater diagnostic
  relevance (\emph{e.g.}, hippocampus in AD).
  We introduce a PACM strategy that biases
  mask sampling toward clinically important regions through a
  temperature-scheduled curriculum.

  \noindent\textbf{Importance Scoring.}
  We register the AAL116 atlas to the input
  space and compute a patch-region membership matrix
  $\mathbf{R} \in \mathbb{R}^{N \times K}$, where $R_{i,k}$ is the
  fraction of voxels in patch $i$ belonging to region $k$.
  A \emph{static} score is derived from neuropathological priors
  (Braak staging~\cite{braak2006staging}):
  $s_i^{\text{s}} = \sum_{k} R_{i,k} \, w_k$,
  where $w_k$ reflects the pathological relevance of region $k$.
  A \emph{dynamic} score $s_i^{\text{d}}$ is obtained by periodically extracting the
  CLS-to-patch attention from an EMA anatomy teacher and aggregating it to
  region level via $\mathbf{R}$.
  The two are combined as
  $s_i = (1{-}\beta)\,\hat{s}_i^{\text{s}} + \beta\,\hat{s}_i^{\text{d}}$
  after min-max normalization.

    \noindent\textbf{Curriculum Schedule.}
  Importance scores are converted to mask probabilities via
  $p_i = \mathrm{softmax}(s_i / \tau)$.
  To avoid early training instability, we adopt a three-phase curriculum:
  (1)~uniform random masking in the early stage;
  (2)~$\tau$ anneals from $\tau_{\text{start}}$ to $\tau_{\text{target}}$
  with cosine decay, progressively sharpening the distribution;
  (3)~$\tau{=}\tau_{\text{target}}$ for the remainder of training.
  Patches are then selected using the Gumbel-top-$k$
  trick~\cite{kool2019stochastic}:
  $k_i = \log p_i + g_i,\; g_i {\sim} \mathrm{Gumbel}(0,1)$,
  where low-importance patches are kept visible and high-importance patches are masked.
  This progressively shifts the model's reconstructive effort toward
  pathologically critical structures.

\subsection{Training Objective and Downstream Finetuning}
\label{sec:finetune}

The overall loss is
$\mathcal{L} = \mathcal{L}_{\text{MAE}}
  + \lambda \cdot \mathbbm{1}[\text{paired}] \cdot \mathcal{L}_{\text{RCMD}}$,
where $\lambda$ follows a linear warmup schedule and
$\mathbbm{1}[\text{paired}]$ activates the cross-modal term only when both
MRI and PET are observed. For finetuning, we only use the student encoder with a lightweight task head (layer normalization + linear layer on the \texttt{[CLS]} token) and remove patch-level masking. Missing modalities remain zeroed and attention-blocked as in pretraining; modality dropout is applied during training for additional robustness.
\section{Experimental Results}
\begin{table*}[t]
\centering
\caption{Modality availability across pretraining datasets and downstream diagnostic classes, reported as count (percentage).}
\label{tab:dataset_modality_stats}
{\renewcommand{\arraystretch}{0.6} 
\setlength{\tabcolsep}{4pt}         
\resizebox{\textwidth}{!}{%
\begin{tabular}{@{} l cccccc @{}}
\toprule
\textbf{Split} & \textbf{Subjects} & \textbf{T1} & \textbf{T2} & \textbf{FLAIR} & \textbf{PET} & \textbf{Total} \\
\midrule
\multicolumn{7}{@{}l}{\textit{Pretraining Datasets}} \\[2pt]
\quad A4      & 1,736  & 1,736 (100.0\%) & 1,719 (99.0\%)  & 1,729 (99.6\%)  & 1,526 (87.9\%) & 6,710  \\
\quad DOD-ADNI & 543   & 499 (91.9\%)   & 0 (0.0\%)      & 0 (0.0\%)      & 284 (52.3\%)  & 783   \\
\quad AIBL    & 1,408  & 1,283 (91.1\%)  & 801 (56.9\%)   & 670 (47.6\%)   & 407 (28.9\%)  & 3,161  \\
\quad BraTS23   & 2,569  & 2,569 (100.0\%) & 2,569 (100.0\%) & 2,569 (100.0\%) & 0 (0.0\%)     & 7,707  \\
\quad NACC    & 7,244  & 7,244 (100.0\%) & 773 (10.7\%)   & 6,637 (91.6\%)  & 1,884 (26.0\%) & 16,538 \\
\cmidrule{2-7}
\quad \textbf{Subtotal} & \textbf{13,500} & \textbf{13,331 (98.7\%)} & \textbf{5,862 (43.4\%)} & \textbf{11,605 (86.0\%)} & \textbf{4,101 (30.4\%)} & \textbf{34,899} \\
\midrule
\multicolumn{7}{@{}l}{\textit{Downstream Dataset (ADNI)}} \\[2pt]
\quad CN  & 1,235 & 1,143 (92.6\%) & 700 (56.7\%) & 1,050 (85.0\%) & 738 (59.8\%) & 3,631 \\
\quad MCI & 1,133 & 1,069 (94.4\%) & 554 (48.9\%) & 869 (76.7\%)  & 496 (43.8\%) & 2,988 \\
\quad AD  & 613  & 573 (93.5\%)  & 276 (45.0\%) & 383 (62.5\%)  & 230 (37.5\%) & 1,462 \\
\cmidrule{2-7}
\quad \textbf{Subtotal} & \textbf{2,981} & \textbf{2,785 (93.4\%)} & \textbf{1,530 (51.3\%)} & \textbf{2,302 (77.2\%)} & \textbf{1,464 (49.1\%)} & \textbf{8,081} \\
\bottomrule
\end{tabular}%
}}
\end{table*}
\begin{table}[t]
\centering
\begingroup
\newcommand{\up}{\ensuremath{\uparrow}}
\newcommand{\down}{\ensuremath{\downarrow}}
\addtolength{\tabcolsep}{0.1em}
\newcommand{\mstd}[2]{#1\,{\scalebox{0.6}{(#2)}}}

\caption{Results on ADNI. The best results are highlighted in \colorbox{pink!38}{pink}, and the second-best results are \underline{underlined}.}
\resizebox{\textwidth}{!}{
\begin{tabular}{
@{\hspace{\origtabcolsep}}
>{\centering\arraybackslash}m{1.6cm}
@{\hspace{0.3em}\vrule width 0.3pt\hspace{0.4em}}
l@{\hspace{\origtabcolsep}}|
ccc|ccc|ccc|ccc}
\toprule
\multirow[c]{2}{*}{Modality}&\multirow[c]{2}{*}{Method}
& \multicolumn{3}{c|}{\textbf{CN vs. AD}}
& \multicolumn{3}{c|}{\textbf{CN vs. MCI}}
& \multicolumn{3}{c|}{\textbf{MMSE prediction}}
& \multicolumn{3}{c}{\textbf{Age prediction (Years)}} \\
\cmidrule{3-14}
& 
& \metbf{ACC\up} & \metbf{AUC\up} & \metbf{F1\up}
& \metbf{ACC\up} & \metbf{AUC\up} & \metbf{F1\up}
& \metbf{MAE \down} & \metbf{RMSE \down} & \metbf{PCC\up}
& \metbf{MAE \down} & \metbf{RMSE \down} & \metbf{PCC\up} \\
\midrule

\multirow[c]{8}{*}{T1} 
& 3D Resnet50~\cite{hara2018can}
& \underline{\mstd{84.2}{2.8}} & \underline{\mstd{93.1}{2.6}} & \underline{\mstd{87.2}{1.6}}
& \mstd{62.6}{4.3} & \underline{\mstd{67.5}{6.2}} & \mstd{72.7}{1.9}
& \underline{\mstd{2.102}{0.116}} & \underline{\mstd{2.730}{0.162}} & \cellcolor{pink!38}\mstd{60.0}{6.1}
& \cellcolor{pink!38}\mstd{3.759}{0.087} & \cellcolor{pink!38}\mstd{4.851}{0.154} & \cellcolor{pink!38}\mstd{78.5}{1.5} \\

& M3T~\cite{jang2022m3t}
& \mstd{70.3}{1.4} & \mstd{81.2}{1.0} & \mstd{75.9}{2.0}
& \mstd{63.4}{3.8} & \mstd{54.7}{4.1} & \underline{\mstd{75.5}{2.4}}
& \mstd{2.370}{0.044} & \mstd{2.964}{0.098} & \mstd{43.5}{2.7}
& \mstd{5.230}{0.082} & \mstd{6.512}{0.067} & \mstd{57.9}{1.0} \\
& BrainIAC~\cite{tak2026generalizable}
& \mstd{76.1}{3.4} & \mstd{82.5}{1.5} & \mstd{79.1}{2.9}
& \mstd{60.9}{2.6} & \mstd{58.4}{0.5} & \mstd{72.0}{1.6}
& \mstd{2.358}{0.084} & \mstd{3.030}{0.160} & \mstd{46.5}{0.9}
& \mstd{5.246}{0.067} & \mstd{6.584}{0.082} & \mstd{53.3}{3.4} \\
& SAM-Brain3D~\cite{DENG2025112595}
& \mstd{81.1}{3.6} & \mstd{91.7}{1.9} & \mstd{84.2}{2.2}
& \mstd{58.9}{1.9} & \mstd{59.3}{2.4} & \mstd{69.8}{2.6}
& \mstd{2.210}{0.106} & \mstd{2.836}{0.144} &\mstd{56.2}{2.6}
& \mstd{4.492}{0.154} & \mstd{5.630}{0.092} & \mstd{68.2}{1.5} \\

& Flex-MoE~\cite{yun2025flex}
& \mstd{75.2}{3.9} & \mstd{81.7}{7.4} & \mstd{81.3}{2.3}
& {\mstd{63.4}{2.9}} & \mstd{59.4}{3.3} & \mstd{74.4}{4.6}
& \mstd{2.660}{0.100} & \mstd{3.560}{0.200} & \mstd{21.7}{1.7}
& \mstd{6.205}{0.667} & \mstd{7.845}{0.923} & \mstd{60.2}{5.6} \\
& FuseMoE~\cite{han2025fusemoe}
& \mstd{74.8}{2.1} & \mstd{80.2}{1.0} & \mstd{80.4}{1.9}
& \mstd{60.9}{3.8} & \mstd{57.8}{1.6} & \mstd{73.8}{4.2}
& \mstd{2.560}{0.040} & \mstd{3.260}{0.160} & \mstd{27.8}{3.0}
& {\mstd{4.871}{0.154}} & \mstd{6.256}{0.154} & \mstd{58.9}{2.8} \\
& MoE-retriever~\cite{yun2025generate}
& \mstd{69.4}{11.3} & \mstd{72.0}{13.9} & \mstd{79.9}{5.4}
& \underline{\mstd{65.8}{0.7}} & \mstd{60.9}{4.5} & \cellcolor{pink!38}{\mstd{79.0}{0.8}}
& \mstd{2.363}{0.187} & \mstd{3.007}{0.178} & \mstd{29.2}{31.6}
& \mstd{5.076}{0.974} & \mstd{6.410}{0.974} & \mstd{62.1}{13.2} \\
& {\textbf{BrainAnytime}}
& \cellcolor{pink!38}\mstd{86.5}{2.7} & \cellcolor{pink!38}\mstd{93.4}{2.4} & \cellcolor{pink!38}\mstd{88.4}{2.9}
& \cellcolor{pink!38}\mstd{66.7}{2.5} & \cellcolor{pink!38}\mstd{70.1}{1.5} & \mstd{75.2}{3.4}
& \cellcolor{pink!38}\mstd{2.100}{0.040} & \cellcolor{pink!38}\mstd{2.660}{0.080} & \underline{\mstd{57.8}{0.4}}
& \underline{\mstd{4.205}{0.051}} & \underline{\mstd{5.435}{0.103}} & \underline{\mstd{71.2}{0.6}} \\

\midrule
\multirow[c]{8}{*}{T1+Flair} & MCAD~\cite{zhang2023multi}
& \mstd{87.8}{3.6} & \mstd{93.9}{1.0} & \mstd{67.9}{10.5}
& \mstd{63.3}{4.2} & \mstd{65.6}{2.9} & \mstd{54.4}{7.0}
& \underline{\mstd{1.910}{0.056}} & \mstd{2.814}{0.130} & \mstd{52.6}{2.5}
& \underline{\mstd{4.205}{0.282}} & \underline{\mstd{5.461}{0.328}} & \underline{\mstd{75.3}{0.8}} \\
& MENet~\cite{leng2023multimodal}
& \mstd{89.6}{2.1} & \mstd{95.4}{0.3} & \mstd{73.5}{5.8}
& \mstd{62.1}{2.9} & \mstd{64.5}{3.6} & \mstd{52.6}{5.4}
& \mstd{1.924}{0.014} & \mstd{2.784}{0.078} & \mstd{53.1}{3.2}
& \cellcolor{pink!38}\mstd{4.102}{0.144} & \cellcolor{pink!38}\mstd{5.317}{0.190} & \cellcolor{pink!38}\mstd{75.8}{0.6} \\
& BrainIAC
& \mstd{80.2}{0.8} & \mstd{87.8}{1.1} & \mstd{58.9}{4.4}
& \mstd{62.1}{0.5} & \mstd{67.6}{3.0} & \mstd{59.0}{1.3}
& \mstd{1.964}{0.062} & \mstd{2.958}{0.092} & \mstd{43.9}{6.3}
& \mstd{5.087}{0.144} & \mstd{6.440}{0.179} & \mstd{63.4}{1.4} \\
& SAM-Brain3D
& \underline{\mstd{91.4}{3.4}} & \mstd{95.3}{1.6} & \cellcolor{pink!38}\mstd{81.9}{6.6}
& \mstd{64.9}{1.4} & \mstd{68.2}{2.2} & \underline{\mstd{62.0}{2.6}}
& \mstd{1.964}{0.038} & \mstd{2.884}{0.052} & \mstd{47.9}{3.1}
& \mstd{4.466}{0.303} & \mstd{5.712}{0.472} & \mstd{73.0}{2.8} \\
& Flex-MoE
& \mstd{88.3}{0.8} & \underline{\mstd{95.5}{0.3}} & \mstd{75.9}{1.8}
& \mstd{58.5}{8.2} & \mstd{62.7}{5.3} & \mstd{50.7}{12.0}
& \mstd{2.320}{0.460} & \mstd{3.140}{0.320} & \mstd{41.7}{8.6}
& \mstd{4.974}{0.615} & \mstd{6.307}{0.513} & \mstd{68.1}{1.7} \\
& FuseMoE
& \mstd{90.1}{0.8} & \mstd{94.3}{0.8} & \mstd{78.3}{2.4}
& \underline{\mstd{65.8}{2.1}} & \underline{\mstd{69.9}{2.3}} & \mstd{61.5}{5.0}
& \mstd{1.980}{0.020} & \mstd{3.000}{0.160} & \mstd{43.0}{4.9}
& \mstd{4.769}{0.103} & \mstd{6.205}{0.154} & \mstd{64.8}{0.9} \\
& MoE-retriever
& \mstd{90.1}{2.1} & \mstd{93.9}{2.6} & \mstd{77.6}{4.2}
& \mstd{53.3}{3.7} & \mstd{57.8}{3.7} & \mstd{58.6}{6.8}
& \cellcolor{pink!38}\mstd{1.878}{0.024} & \underline{\mstd{2.748}{0.165}} & \cellcolor{pink!38}\mstd{57.1}{2.4}
& \mstd{4.666}{0.308} & \mstd{5.948}{0.308} & \mstd{71.1}{4.8} \\
& \textbf{BrainAnytime}
& \cellcolor{pink!38}\mstd{91.9}{2.7} & \cellcolor{pink!38}\mstd{97.0}{1.6} & \underline{\mstd{81.3}{5.6}}
& \cellcolor{pink!38}\mstd{73.3}{1.4} & \cellcolor{pink!38}\mstd{78.8}{2.2} & \cellcolor{pink!38}\mstd{70.0}{3.9}
& \mstd{1.920}{0.060} & \cellcolor{pink!38}\mstd{2.700}{0.100} & \underline{\mstd{55.9}{4.9}}
& \mstd{4.256}{0.051} & \mstd{5.538}{0.051} & \mstd{74.2}{1.9} \\

\midrule
\multirow[c]{4}{*}{\makecell[c]{T1+T2+\\Flair}} & BrainIAC
& \mstd{76.6}{2.3} & \mstd{82.1}{1.4} & \mstd{62.8}{3.6}
& \mstd{54.5}{2.8} & \mstd{56.2}{1.4} & \mstd{54.0}{4.3}
& \mstd{2.578}{0.040} & \mstd{3.666}{0.068} & \mstd{35.4}{4.2}
& \mstd{4.953}{0.118} & \mstd{6.246}{0.046} & \mstd{67.8}{1.1} \\
& SAM-Brain3D
& \mstd{76.6}{5.6} & \mstd{85.8}{4.6} & \mstd{64.2}{7.5}
& \cellcolor{pink!38}\mstd{59.9}{1.2} & \underline{\mstd{63.4}{1.8}} & \cellcolor{pink!38}\mstd{59.5}{3.3}
& \mstd{2.414}{0.158} & \mstd{3.292}{0.274} & \mstd{52.0}{7.6}
& \underline{\mstd{3.953}{0.174}} & \underline{\mstd{4.953}{0.205}} & \underline{\mstd{80.3}{2.0}} \\
& Flex-MoE
& \mstd{77.7}{3.2} & \underline{\mstd{88.8}{1.5}} & \mstd{61.8}{7.4}
& \mstd{47.7}{3.1} & \mstd{61.6}{2.3} & \mstd{12.7}{12.6}
& \mstd{2.500}{0.080} & \mstd{3.400}{0.240} & \mstd{47.7}{3.5}
& \mstd{4.666}{0.769} & \mstd{5.743}{0.820} & \mstd{77.5}{3.3} \\
& FuseMoE
& \mstd{76.9}{2.2} & \mstd{83.0}{0.7} & \mstd{60.5}{6.9}
& \mstd{55.3}{2.1} & \mstd{58.6}{2.5} & \mstd{44.5}{0.4}
& \mstd{2.960}{0.180} & \mstd{4.120}{0.220} & \mstd{38.4}{6.2}
& \mstd{4.564}{0.051} & \mstd{5.846}{0.051} & \mstd{73.1}{0.9} \\
& MoE-retriever
& \underline{\mstd{82.1}{5.1}} & \mstd{86.9}{6.2} & \underline{\mstd{72.9}{7.9}}
& \mstd{55.3}{2.2} & \mstd{62.5}{4.5} & \underline{\mstd{57.8}{12.2}}
& \underline{\mstd{2.301}{0.062}} & \cellcolor{pink!38}\mstd{3.075}{0.209} & \underline{\mstd{58.3}{3.1}}
& \mstd{4.205}{0.308} & \mstd{5.282}{0.308} & \mstd{77.3}{4.2} \\
& \textbf{BrainAnytime}
& \cellcolor{pink!38}\mstd{84.2}{1.7} & \cellcolor{pink!38}\mstd{90.6}{0.5} & \cellcolor{pink!38}\mstd{77.0}{5.6}
& \underline{\mstd{59.6}{2.6}} & \cellcolor{pink!38}\mstd{65.4}{3.5} & \mstd{54.0}{8.4}
& \cellcolor{pink!38}\mstd{2.260}{0.040} & \underline{\mstd{3.140}{0.060}} & \cellcolor{pink!38}\mstd{63.3}{3.3}
& \cellcolor{pink!38}\mstd{3.795}{0.256} & \cellcolor{pink!38}\mstd{4.871}{0.256} & \cellcolor{pink!38}\mstd{82.6}{0.4} \\

\midrule
\multirow[c]{4}{*}{\makecell[c]{T1+Flair+\\PET}}  & Flex-MoE
& \underline{\mstd{90.4}{1.8}} & \underline{\mstd{95.3}{0.3}} & \underline{\mstd{84.1}{2.7}}
& \mstd{59.6}{5.6} & \mstd{63.8}{11.9} & \mstd{33.9}{20.5}
& \mstd{2.180}{0.180}& \mstd{2.960}{0.080} & \mstd{46.9}{1.0}
& \mstd{4.974}{0.205} & \mstd{6.205}{0.308} & \mstd{70.5}{2.2} \\
& FuseMoE
& \mstd{82.3}{2.3} & \mstd{92.0}{1.9} & \mstd{63.7}{6.4}
& \underline{\mstd{66.3}{1.7}} & \underline{\mstd{77.2}{1.0}} & \mstd{55.9}{3.9}
& \mstd{2.220}{0.220} & \mstd{3.220}{0.260} & \mstd{39.0}{6.5}
& \mstd{4.871}{0.051} & \mstd{6.102}{0.103} & \mstd{66.2}{1.2} \\
& MoE-retriever
& \mstd{84.8}{1.5} & \mstd{92.5}{0.8} & \mstd{73.1}{4.2}
& \mstd{65.9}{1.7} & \mstd{72.3}{5.2} & \underline{\mstd{58.1}{5.3}}
& \underline{\mstd{1.956}{0.085}} & \underline{\mstd{2.802}{0.095}} & \underline{\mstd{54.2}{2.3}}
& \underline{\mstd{4.410}{0.359}} & \underline{\mstd{5.640}{0.461}} & \underline{\mstd{71.3}{4.0}} \\
& \textbf{BrainAnytime}
& \cellcolor{pink!38}\mstd{92.4}{1.5} & \cellcolor{pink!38}\mstd{97.5}{0.8} & \cellcolor{pink!38}\mstd{87.2}{2.7}
& \cellcolor{pink!38}\mstd{68.9}{2.9} & \cellcolor{pink!38}\mstd{77.8}{1.4} & \cellcolor{pink!38}\mstd{66.2}{1.6}
& \cellcolor{pink!38}\mstd{1.840}{0.060} & \cellcolor{pink!38}\mstd{2.660}{0.100} & \cellcolor{pink!38}\mstd{59.4}{2.3}
& \cellcolor{pink!38}\mstd{4.051}{0.154} & \cellcolor{pink!38}\mstd{5.128}{0.205} & \cellcolor{pink!38}\mstd{78.6}{1.8} \\

\midrule
\multirow[c]{4}{*}{\makecell[c]{T1+T2+\\Flair+PET}}  & Flex-MoE
& \underline{\mstd{89.3}{1.8}} & \underline{\mstd{93.6}{0.2}} & \underline{\mstd{70.7}{3.3}}
& \mstd{65.3}{2.7} & \mstd{68.3}{1.0} & \mstd{30.7}{12.5}
& \mstd{2.600}{0.500} & \mstd{3.460}{0.640} & \mstd{54.1}{3.1}
& \mstd{4.256}{0.103} & \mstd{5.384}{0.205} & \mstd{75.4}{0.6} \\
& FuseMoE
& \mstd{86.3}{0.6} & \mstd{87.9}{0.4} & \mstd{56.0}{1.6}
& \underline{\mstd{65.9}{1.7}} & \cellcolor{pink!38}\mstd{73.5}{1.0} & \mstd{35.1}{3.2}
& \mstd{2.260}{0.340} & \mstd{3.320}{0.320} & \mstd{45.2}{7.3}
& \mstd{4.564}{0.051} & \mstd{5.794}{0.103} & \mstd{66.3}{1.0} \\
& MoE-retriever
& \mstd{89.1}{3.6} & \mstd{90.6}{6.3} & \mstd{69.9}{10.0}
& \mstd{62.8}{2.3} & \mstd{64.7}{2.6} & \underline{\mstd{38.9}{3.6}}
& \underline{\mstd{1.828}{0.041}} & \underline{\mstd{2.710}{0.077}} & \underline{\mstd{62.9}{1.4}}
& \underline{\mstd{4.153}{0.410}} & \underline{\mstd{5.128}{0.461}} & \underline{\mstd{75.8}{4.2}} \\
& \textbf{BrainAnytime}
& \cellcolor{pink!38}\mstd{91.9}{0.9} & \cellcolor{pink!38}\mstd{95.2}{0.6} & \cellcolor{pink!38}\mstd{78.3}{1.6}
& \cellcolor{pink!38}\mstd{67.3}{2.6} & \underline{\mstd{70.3}{0.8}} & \cellcolor{pink!38}\mstd{54.2}{5.6}
& \cellcolor{pink!38}\mstd{1.800}{0.080} & \cellcolor{pink!38}\mstd{2.680}{0.120} & \cellcolor{pink!38}\mstd{64.5}{2.1}
& \cellcolor{pink!38}\mstd{3.743}{0.154} & \cellcolor{pink!38}\mstd{4.820}{0.154} & \cellcolor{pink!38}\mstd{79.6}{0.6} \\

\midrule
\textbf{\multirow[c]{4}{*}{Average}}& Flex-MoE
& \underline{\mstd{84.2}{2.3}} & \underline{\mstd{91.0}{1.9}} & \underline{\mstd{74.8}{3.5}}
& \mstd{58.9}{4.5} & \mstd{63.2}{4.8} & \mstd{40.5}{12.4}
& \mstd{2.452}{0.260} & \mstd{3.304}{0.300} &\mstd{42.4}{3.6}
& \mstd{5.015}{0.461} & \mstd{6.297}{0.564} & \mstd{70.3}{2.7} \\
& FuseMoE
& \mstd{82.1}{1.6} & \mstd{87.5}{0.9} & \mstd{67.8}{3.8}
& \underline{\mstd{62.8}{2.3}} & \underline{\mstd{67.4}{1.7}} & \mstd{54.2}{3.3}
& \mstd{2.400}{0.160} & \mstd{3.384}{0.220} & \mstd{38.7}{5.6}
& \mstd{4.728}{0.103} & \mstd{6.041}{0.103} & \mstd{65.9}{1.4} \\
& MoE-retriever
& \mstd{83.1}{4.7} & \mstd{87.2}{6.0} & \mstd{74.7}{6.3}
& \mstd{60.6}{2.1} & \mstd{63.6}{4.1} & \underline{\mstd{58.5}{5.5}}
& \underline{\mstd{2.065}{0.053}} & \underline{\mstd{2.868}{0.093}} & \underline{\mstd{52.3}{6.8}}
& \underline{\mstd{4.502}{0.461}} & \underline{\mstd{5.682}{0.513}} & \underline{\mstd{71.5}{6.1}} \\
& \textbf{BrainAnytime}
& \cellcolor{pink!38}\mstd{89.4}{1.9} & \cellcolor{pink!38}\mstd{94.7}{1.2} & \cellcolor{pink!38}\mstd{82.4}{3.7}
& \cellcolor{pink!38}\mstd{67.2}{2.4} & \cellcolor{pink!38}\mstd{72.5}{1.9} & \cellcolor{pink!38}\mstd{63.9}{4.6}
& \cellcolor{pink!38}\mstd{1.984}{0.056} & \cellcolor{pink!38}\mstd{2.768}{0.092} & \cellcolor{pink!38}\mstd{60.2}{2.6}
& \cellcolor{pink!38}\mstd{4.010}{0.133} & \cellcolor{pink!38}\mstd{5.158}{0.154} & \cellcolor{pink!38}\mstd{77.2}{1.1} \\

\bottomrule
\end{tabular}%
}
\label{tab:adni_all_metrics_best_second}
\endgroup
\end{table}
\subsection{Dataset Description}
\label{sec:dataset}
We curate a multi-cohort neuroimaging collection from six publicly available datasets totaling 16,481 subjects (Tab.~\ref{tab:dataset_modality_stats}). Five datasets (A4~\cite{sperling2020association}, DOD-ADNI~\cite{weiner2017effects}, AIBL~\cite{ellis2009australian}, BraTS23, and NACC~\cite{beekly2007national}; 13,500 subjects, 34,899 scans) spanning multi-sequence MRI and amyloid-PET are used for pretraining; an independent ADNI cohort~\cite{weiner2013alzheimer} (2,981 subjects) is held out for evaluation. We designate five clinically common modality combinations (T1; T1+FLAIR; T1+FLAIR+PET; T1+T2+FLAIR; T1+T2+FLAIR+PET) for test reporting with a diagnosis-stratified train/val/test split (60/10/30); remaining low-frequency combinations are used for train/val only (80/20). All volumes undergo skull stripping, affine registration to MNI152, min--max normalization, and resampling to $128\times128\times128$. We evaluate on four downstream tasks~\cite{ding2025denseformer, tak2026generalizable}: CN vs.\ MCI, CN vs.\ AD, MMSE regression, and age prediction.

\subsection{Implementation Details}
The encoder is ViT-B ($d{=}768$, 12 heads, 12 layers, patch size $p{=}16$, $N{=}512$ patches); each decoder has 2 blocks with dimension 384.
We pretrain with AdamW (lr $10^{-4}$, weight decay $0.05$) for 1000 epochs (40-epoch warmup, cosine decay) on 8$\times$A100 GPUs, batch size 128.
Mask ratio $r{=}0.75$, Dirichlet $\alpha{=}1.0$; $\lambda{=}0.1$; $\mu_0{=}0.996$.
PACM uses importance weights $w_k \in \{3.0, 1.5, 0.3\}$ for AD-critical, other gray-matter, and non-brain regions ($\beta{=}0.5$), with curriculum phases at 20\% and 70\% of training ($\tau$: $5.0 {\to} 1.0$).
Data augmentation includes random flipping, affine transforms, and elastic deformation.
For finetuning, we use AdamW (lr $10^{-5}$, weight decay $0.05$) for up to 100 epochs with early stopping (patience 15); the encoder is frozen for the first 5 epochs. Classification uses BCE; regression uses MSE.

\subsection{Benchmarking with Clinically Driven Modality Escalation}
\begin{figure}[!h]
\centerline{\includegraphics[width=0.9\columnwidth]{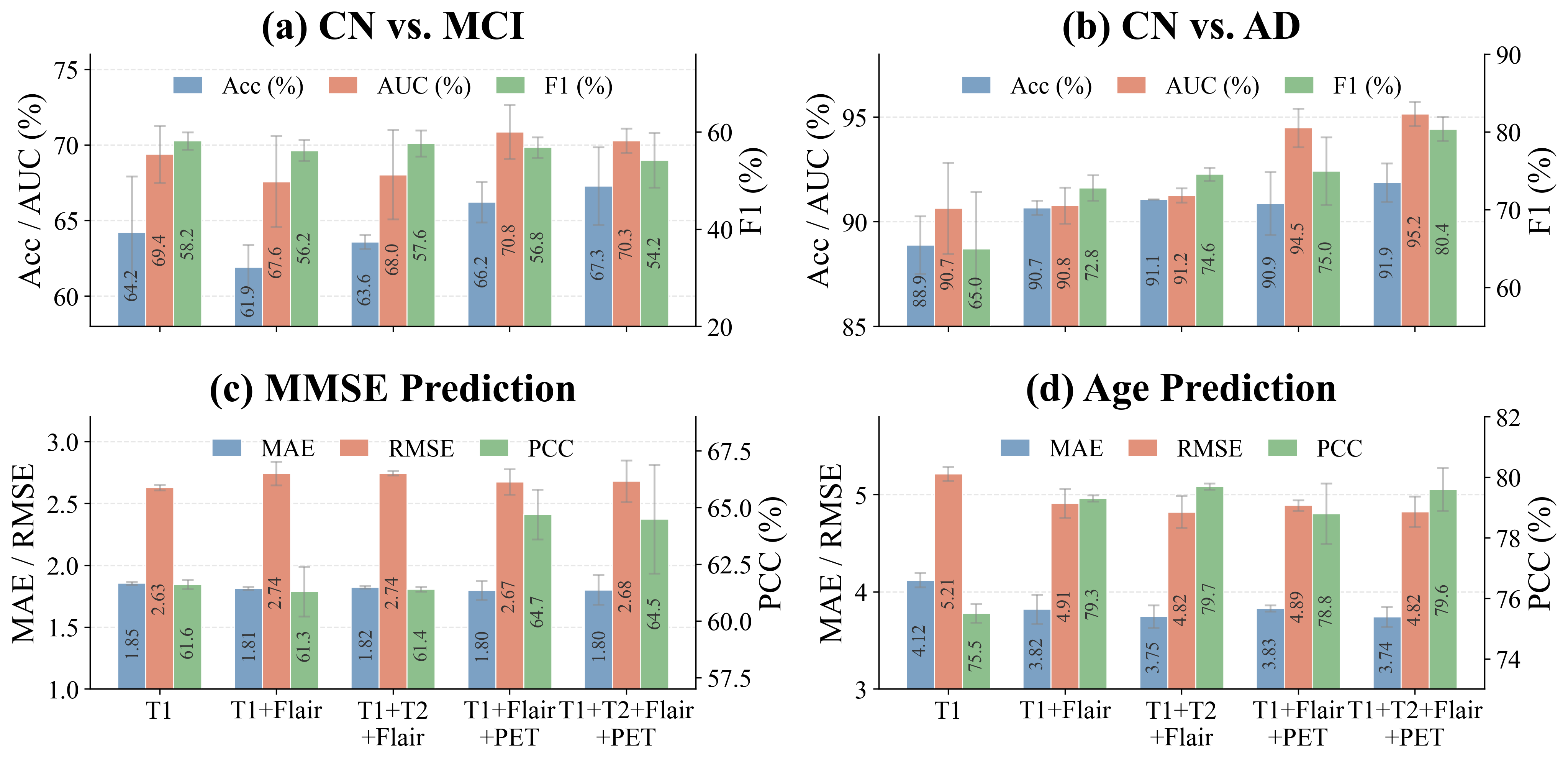}}

\caption{Modality robustness analysis across four downstream tasks under simulated missing-modality conditions via artificial modality dropout.} 
\label{fig:robutness}
\end{figure}

To benchmark BrainAnytime under clinically realistic, stage-dependent modality availability, we compare it with four families of baselines: (i) single-modality models (3D ResNet~\cite{hara2018can}, M3T~\cite{jang2022m3t}), (ii) dual-modality fusion methods (MCAD~\cite{zhang2023multi}, MENet~\cite{leng2023multimodal}), (iii) brain foundation models (BrainIAC~\cite{tak2026generalizable}, SAM-Brain3D~\cite{DENG2025112595}), and (iv) incomplete multimodal learning approaches (Flex-MoE~\cite{yun2025flex}, FuseMoE~\cite{han2025fusemoe}, MoE-retriever~\cite{yun2025generate}). Following the escalation-motivated modality patterns in our cohort (Sec.~\ref{sec:dataset}), we report results on five mutually exclusive modality combinations. For classification, we use ACC/AUC/F1; for MMSE and age regression, we report MAE/RMSE/PCC~\cite{ding2025denseformer}. All results are averaged over 3 seeds (mean(std)). As shown in Table~\ref{tab:adni_all_metrics_best_second}, BrainAnytime achieves the best overall performance across four downstream tasks and five clinically motivated modality settings, with the highest average ACC/AUC/F1 of 89.4/94.7/82.4 for AD diagnosis, outperforming incomplete multimodal learning baselines. Across settings, BrainAnytime ranks first or second on the majority of primary metrics for classification (AUC) and regression (MAE), demonstrating strong effectiveness under arbitrary modality availability. Because the five real-world modality groups in Table~\ref{tab:adni_all_metrics_best_second} are disjoint, cross-group differences may reflect cohort variation rather than modality effects. We thus conduct a within-subject robustness study on the fully observed (T1+T2+FLAIR+PET) test subset, where we synthetically drop modalities to recreate the same five combinations on identical subjects. As additional modalities are provided, performance improves from T1 to full input across all tasks (CN vs.\ AD F1: 65.0$\rightarrow$80.4; CN vs.\ MCI ACC: 64.2$\rightarrow$67.3; MMSE PCC: 61.6$\rightarrow$64.5; Age PCC: 75.5$\rightarrow$79.6; Fig.~\ref{fig:robutness}).
\begin{table*}[!h]
\centering
\newcommand{\mstd}[2]{#1\,{\scalebox{0.8}{(#2)}}}
\newcommand{\best}[1]{\cellcolor{pink!38}#1}
\newcommand{\second}[1]{\underline{#1}}

\begin{minipage}[t]{0.55\linewidth}
  \centering
  \captionsetup{type=table}
  \caption{Ablation study}
  \label{tab:ablation_merged_highlighted}
  \setlength{\tabcolsep}{4pt}
  \resizebox{\linewidth}{!}{%
  \begin{tabular}{clccc}
    \toprule
    Task & Method & ACC & AUC & F1 \\
    \midrule
    \multirow{5}{*}{\rotatebox{90}{CN vs.\ AD}}
    & ViT-B (scratch)      & \mstd{78.8}{2.2} & \mstd{84.8}{1.1} & \mstd{63.3}{5.7} \\
    & Multi-MAE3D           & \mstd{86.9}{0.4} & \mstd{92.8}{1.9} & \mstd{75.6}{1.5} \\
    & Multi-MAE3D+RCMD      & \second{\mstd{88.8}{0.3}} & \mstd{94.1}{0.9} & \second{\mstd{81.3}{0.8}} \\
    & Multi-MAE3D+PACM      & \mstd{88.1}{1.6} & \best{\mstd{94.8}{0.5}} & \mstd{79.9}{3.7} \\
    & BrainAnytime (Ours)  & \best{\mstd{89.4}{0.5}} & \second{\mstd{94.7}{0.5}} & \best{\mstd{82.7}{1.5}} \\
    \midrule
    \multirow{5}{*}{\rotatebox{90}{CN vs.\ MCI}}
    & ViT-B (scratch)      & \mstd{59.3}{2.1} & \mstd{61.4}{1.4} & \mstd{48.6}{9.2} \\
    & Multi-MAE3D           & \mstd{64.4}{1.8} & \mstd{68.1}{1.7} & \mstd{61.2}{3.8} \\
    & Multi-MAE3D+RCMD      & \mstd{65.7}{0.6} & \mstd{71.3}{0.2} & \best{\mstd{64.2}{2.0}} \\
    & Multi-MAE3D+PACM      & \second{\mstd{66.7}{1.1}} & \second{\mstd{71.5}{1.1}} & \mstd{61.2}{2.4} \\
    & BrainAnytime (Ours)  & \best{\mstd{67.2}{1.4}} & \best{\mstd{72.5}{0.9}} & \second{\mstd{63.9}{3.6}} \\
    \bottomrule
  \end{tabular}%
  }
\end{minipage}
\hfill
\begin{minipage}[t]{0.39\linewidth}
  \centering
  \captionsetup{type=figure}
  \caption{(a) Dynamic score in PACM and (b) attention map.}
  \label{fig:attention}
  \includegraphics[width=\linewidth]{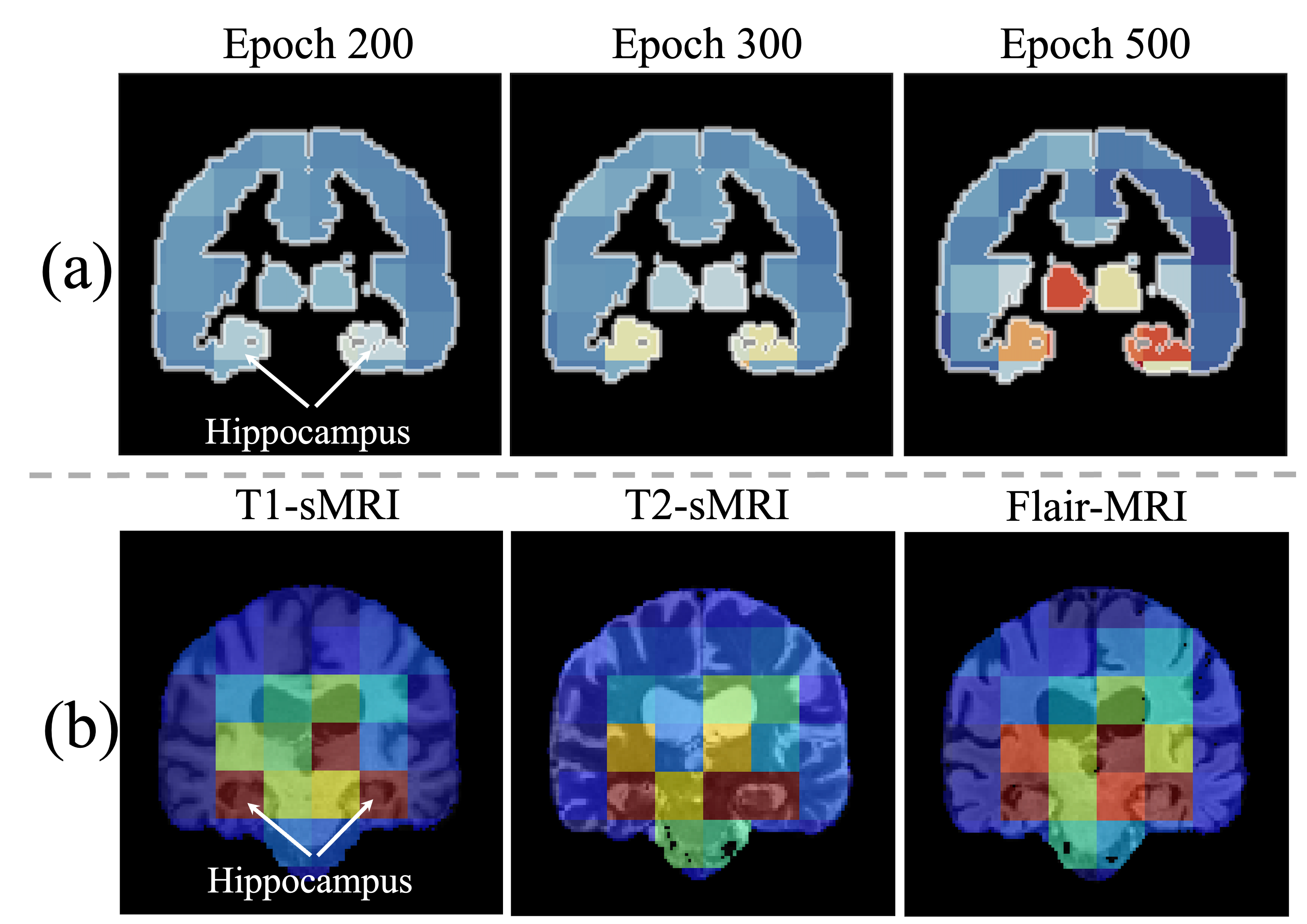}
\end{minipage}
\end{table*}

\subsection{Ablation Study and Model Interpretability}
\begin{figure}[!t]
\centerline{\includegraphics[width=0.9\columnwidth]{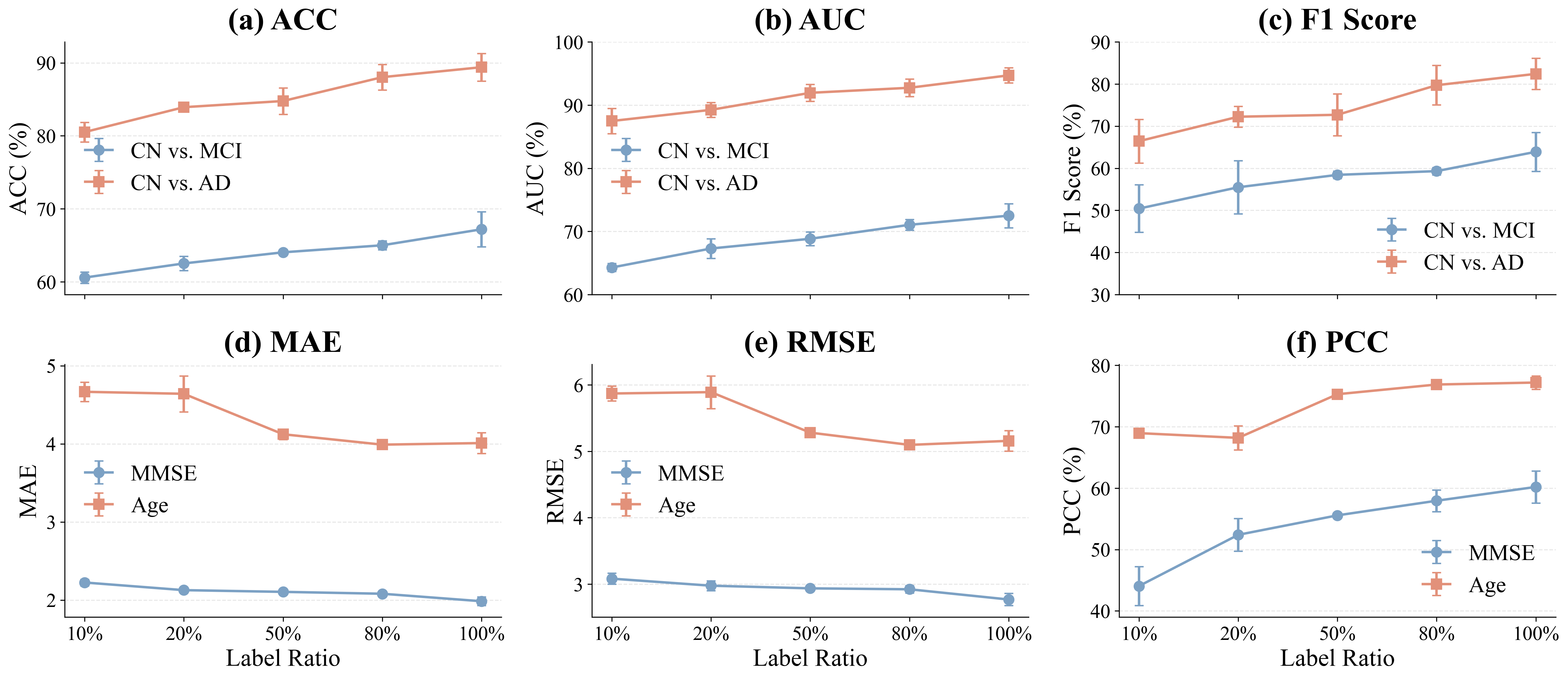}}

\caption{Label efficiency analysis of BrainAnytime with 10\%, 20\%, 50\%, 80\%, and 100\% of training data on downstream tasks.} 

\label{fig:label_efficiency}
\end{figure}

Table~\ref{tab:ablation_merged_highlighted} ablates each component. MultiMAE3D pretraining already yields a large gain over training from scratch; adding RCMD or PACM individually further improves results, while combining both (BrainAnytime) achieves the best ACC/F1 on CN vs.\ AD (89.4/82.7) and ACC/AUC on CN vs.\ MCI (67.2/72.5) averaged cross all five modality settings. Label efficiency analysis (Fig.~\ref{fig:label_efficiency}) confirms reduced reliance on labeled data. We also visualize the PACM dynamic score evolution (Fig.~\ref{fig:attention}(a)) and the attention map of a predicted AD sample (Fig.~\ref{fig:attention}(b)), showing that BrainAnytime attends to AD-relevant regions such as the hippocampus.

\section{Conclusion}
We presented BrainAnytime, a unified framework pretrained on 34,899 3D brain scans
that supports brain image analysis under arbitrary modality availability.
Through cross-modal distillation and atlas-guided curriculum masking,
a single model flexibly accepts whatever imaging is available and
consistently outperforms modality-specific, missing-modality,
and foundation model baselines across four tasks and five modality settings.

\medskip\noindent\textbf{Acknowledgments.} 
\small 
This work was partially supported by RGC Collaborative Research Fund (No. C5055-24G), the Start-up Fund of The Hong Kong Polytechnic University (No. P0045999), the Seed Fund of the Research Institute for Smart Ageing (No. P0050946), and Tsinghua-PolyU Joint Research Initiative Fund (No. P0056509), and PolyU UGC funding (No. P0053716).

%
%
%
%
\bibliographystyle{splncs04}
\bibliography{main}

\end{document}